\documentclass[10pt, conference]{IEEEtran}
\IEEEoverridecommandlockouts
\usepackage{cite}
\usepackage{amsmath,amssymb,amsfonts}
\usepackage{algorithmic}
\usepackage{graphicx}
\usepackage{textcomp}
\usepackage{xcolor}
\usepackage{enumerate}

\usepackage{subcaption}
\usepackage{url}
\usepackage{hhline}
\usepackage{hyperref}
\usepackage{dsfont}
\usepackage{amsthm}
\usepackage{verbatim}
\usepackage[linesnumbered,ruled,vlined]{algorithm2e}

\usepackage{threeparttable,caption,tabularx,booktabs}
\newcolumntype{L}{>{\raggedright\arraybackslash}X}
\newcolumntype{R}{>{\raggedleft\arraybackslash}X}

\newcommand{\figlabel}[1]{\label{#1}}
\newcommand{\seclabel}[1]{\label{#1}}
\newcommand{\tablabel}[1]{\label{#1}}
\newcommand{\secref}[1]{\autoref{#1}}  
\newcommand{\figref}[1]{\autoref{#1}}  
\newcommand{\tabref}[1]{\autoref{#1}}  

\newtheorem{mydef}{Definition}

\let\originallesssim\lesssim
\let\originalgtrsim\gtrsim

\DeclareRobustCommand{\lesssim}{%
  \mathrel{\mathpalette\lowersim\originallesssim}%
}
\DeclareRobustCommand{\gtrsim}{%
  \mathrel{\mathpalette\lowersim\originalgtrsim}%
}

\makeatletter
\newcommand{\lowersim}[2]{%
  \sbox\z@{$#1<$}%
  \raisebox{-\dimexpr\height-\ht\z@}{$\m@th#1#2$}%
}
\makeatother

\def\BibTeX{{\rm B\kern-.05em{\sc i\kern-.025em b}\kern-.08emT\kern-.1667em\lower.7ex\hbox{E}\kern-.125emX}}

\begin{document}

\title{Sampling for Deep Learning Model Diagnosis \\ {\large TECHNICAL REPORT}}

\author{\IEEEauthorblockN{Parmita Mehta\IEEEauthorrefmark{1}, Stephen Portillo\IEEEauthorrefmark{2},
Magdalena Balazinska\IEEEauthorrefmark{1}, Andrew Connolly\IEEEauthorrefmark{2}}
\IEEEauthorblockA{University Of Washington\\
\IEEEauthorrefmark{1}Paul Allen School of Computer Science and Engineering,
\IEEEauthorrefmark{2}DiRAC Institute, Department of Astronomy}}
\maketitle

\begin{abstract}

Deep learning (DL) models have achieved paradigm-changing performance in
many fields with high dimensional data, such as images, audio, and
text. However, the black-box nature of deep neural networks is a barrier not
just to adoption in applications such as medical diagnosis, where interpretability is
essential, but also impedes diagnosis of under performing models. The task
of diagnosing or explaining DL models
requires the computation of additional artifacts, such as activation
values and gradients.  These artifacts are large in volume, and their
computation, storage, and querying raise significant data management
challenges.

In this paper, we articulate DL diagnosis as a data
management problem, and we propose a general, yet representative, set of
queries to evaluate systems that strive to support this new
workload. We further develop a novel data sampling technique that
produce approximate but accurate results for these model debugging
queries. Our sampling technique utilizes the lower dimension
representation learned by the DL model and focuses on model
decision boundaries for the data in this lower dimensional space. We
evaluate our techniques on one standard computer vision and one
scientific data set and demonstrate that our sampling technique
outperforms a variety of state-of-the-art alternatives in terms of
query accuracy.




\end{abstract}
\maketitle

\begin{sloppypar}

\section{Introduction}
\seclabel{intro}

Deep learning (DL) models have enabled unprecedented breakthroughs in developing artificial intelligence systems for analyzing
high-dimensional data, such as text, audio, and images. Building such models is a \textit{data intensive} task.
To build an effective model, a machine learning (ML) practitioner needs to proceed in an iterative fashion, building and tuning dozens of models before selecting one.
While naive selection of the \textit{best} model could be based on statistical measures such as, accuracy, F1 score, etc.,
examining what the model is learning and why it is making mistakes requires access to artifacts, such as model activations and gradients.
Activation values, or \textit{activations}, are learned representations of input data. \textit{Gradients} are partial derivatives of the target output
(e.g., the true label of the input data) with respect to the input data. At a high level, activations and gradients are high
dimensional vectors with sizes that depend on input data dimensionality and DL model architecture. While activations
depict what the deep learning model \textit{sees}, gradients depict \textit{areas of high model sensitivity}.

The naive solution of pre-computing and storing all artifacts required for model diagnosis scales as the product of size
of input data and number of parameters of the deep learning model. For instance, consider a VGG-16~\cite{vgg} model trained on CIFAR-10~\cite{cifar}.
CIFAR-10 is a moderate sized data set with 60k images, $32x32x3$ pixels each;  VGG-16~\cite{vgg} is a deep learning model
with 22 layers and $33,638,218$ learned parameters. Storing activations for ten experiments of training CIFAR-10 data on a
VGG-16 models results in 350GB of data. Although the total number of artifacts for small data sets and models is manageable,
an overhead that is three orders of magnitude larger than the input data per model is not scalable. This makes it difficult to
efficiently perform diagnosis tasks, often preventing interactive diagnosis. Thus, with hundreds of gigabytes of
artifacts per model, building, diagnosing, and selecting a DL model becomes a large-scale data management challenge.

Previous attempts at solving this problem either pre-generated all data required to provide interactive query
times~\cite{kahng16,kahng18, Liu17, Strobelt18} or utilized a variety of storage optimization techniques to manage the storage footprint~\cite{Vartak18, Miao17}.
Both approaches require pre-generated artifacts. Several visualization
tools pre-generate some of the aggregates and severely limit the type of queries that can be posed, while others simply do the latter.
Systems with storage optimizations ~\cite{Vartak18} reduce the storage required for this data by utilizing techniques
such as de-duplication and quantization, etc.

Sampling is a fast and flexible database technique for approximate query processing,
it works well in high dimensions~\cite{Agarwal13,Acharya99, Babcock03, Hellerstein97} and is a potential candidate for this workload.
However, many queries posed for model diagnosis depend on retrieving the top-k maximally activated neuron(s) (see \secref{workload} for workload characterization).
Processing these queries from samples is difficult. The natural estimator for a top-k query over the
sample is the top-k on the sample; however, to un-bias this sample we need to access the full
distribution of frequencies in the un-sampled data set, which is the set of activations for the entire data set over the entire model,
a number far too vast ot generate or store.

The key insight we discovered for creating a sample that can be utilized for DL model diagnosis is that
 the DL model, along with its other objectives such as classification,
learns a \textit{lower dimensional representation} of the data. DL training transforms the input data, creating a new representation with each
layer.  Therefore, to diagnose the model,
we leverage this lower dimensional representation of data rather than store and analyze the distribution of activation values to
create a sample (see \secref{sys} for details). The sampling technique that we develop specifically targets model diagnosis queries, which
include top-k queries as well as average values and, provide more accurate answers than uniform sampling, stratified sampling, and other sampling techniques from the literature.
Our technique selects a sample from the input (test and training) data set so that artifacts need to be generated
only for the sample. This approach not only reduces the storage footprint and speeds-up queries since we store less data, but it also
speeds-up the overall diagnosis process by saving the time it would otherwise take to generate all artifacts for the entire data set.

%

In summary, our contributions include:
\begin{itemize}
\item Characterizing requirements of DL model diagnosis by
   studying debugging queries in the literature. We further develop a simple benchmark for this
  novel workload by generalizing individual queries used in model debugging papers into query sets that cover
  families of queries (\secref{workload}).
\item Presenting a new technique for creating samples for DL model diagnosis (\secref{sys}).
\item Evaluating our approach on two data sets and demonstrating its performance compared with a variety of state-of-the-art alternatives.
\end{itemize}

Our sampling technique can be used to debug any deep learning
model where a lower dimensional representation of the input data is learned in a supervised, semi-supervised or
unsupervised manner.


\section{Preliminaries}
\seclabel{bground}

%
We now summarize current approaches for
DL model diagnosis and their associated data management
challenges, which we address in this paper.

A DL model takes as input a vector $x = [x_{1},\dots, x_{N}],\in
\mathds{R}^{N}$
and produces as output another vector $S(x) = [S_{1}(x), \dots, S_{C}(x)], $ where $C$ is the total number of output
neurons. DL models are constructed in layers, intermediate layers are called \textit{hidden layers}, and each hidden layer
of the model is vector-valued. The dimensionality of these hidden layers determines the width of the model,
and the number of hidden layers determines its depth. These layers often
perform different operations such as convolutions, pooling, dropout, etc. -
and are named accordingly. When the model is evaluated over an input data
instance, such as an image, it produces a value for each $C$ neuron.
The raw values thus produced are activations, and derivatives of these values with respect to a
target, such as class label, are gradients. Diagnosis of DL models relies on these artifacts. The ML community has
a variety of techniques to diagnose these models, which we discuss below.

\subsubsection{Visualization} Manual visual inspection, is a popular diagnosis technique for DL models~\cite{ girshick14, karpathy15, kahng16, kahng18,Liu17}. Standalone tools for visual inspection of DL models built on image data (Cnnvis~\cite{Liu17}) and text data (Activis~\cite{kahng18}, LSTMvis~\cite{Strobelt18}) have been proposed. Some visualization
capability is also integrated with deep learning libraries (e.g. Tensorboard~\cite{tensorflow}, etc.).
These tools provide static and interactive
visualizations of DL model activations and on occasion, gradients. They let ML practitioners
view activation or gradient patterns for various layers as well as view aggregates (e.g., average activation) over
sets of input data instances belonging to each class, which may be classified correctly or incorrectly. This lets ML
practitioners identify specific neuron pattern anomalies and neuron groups and data instances that
require further investigation. \textbf{Challenge:} The size of the artifacts required for these visualizations depends on the size of the input data, and the complexity of the model.
It can easily be $3$ orders of magnitude larger than the input
data set as shown in \tabref{tab:data}. To support interactive visualization for
arbitrary queries, these artifacts must be pre-computed since real-time
computation is too slow to be interactive. To
deal with the associated data explosion, tools such as Activis~\cite{kahng18} limit
the number of layers that can be visualized in the tool.

\begin{table}[t]
    \centering
\setlength\tabcolsep{3pt}
  \begin{tabular}{p{2 cm} |p{1.5cm }|p{1.5 cm}|p{2 cm }}
  \toprule
    \textbf{Data set name} & \textbf{Image size (KB)} & \textbf{Number of model parameters} &
    \textbf{Ratio of size of activations per image} \\
    \midrule
    MNIST & 0.78 &  107,786 & 53 $\times$ \\
    Galaxy Zoo2  & 1.3  & 1,095,842 & 2905.5 $\times$ \\
  \bottomrule
  \end{tabular}
    \caption{Data size and model sizes for standard ML data set (MNIST) and
    scientific image data set (Galaxy Zoo2).}
    \tablabel{tab:data}
\vspace{-2.0em}
\end{table}

\subsubsection {Examining learned representation} A DL model simultaneously learns a lower level representation
of the data and a classifier (in the case of supervised learning). The
learned representation (activations of neurons over an input data set)
 is used for a variety of goals, such as understanding how a model discriminates between different classes,
 comparing different model architectures or hyper-parameters, and examining how learning progresses over time by
 analyzing representations at various checkpoints in the learning process~\cite{raghu17, Morcos18, kornblith19}. \textbf{Challenge:} These analyses
require activations for the \textit{entire} model(s) over the \textit{entire} input data set. If the training process is being examined, the activations
for multiple checkpoints must be generated and stored. As above,
the required artifacts, especially if diagnosing multiple models or
multiple checkpoints, can result in a data explosion.

\subsubsection{Feature visualization and saliency analysis} The feature visualization techniques answer questions about what a DL model or parts thereof
are looking for by generating examples from the learned model~\cite{olah17}. \textit{Feature visualization} uses derivatives to iteratively modify
an input, such as random noise, with the goal of finding the input that maximally activates a particular neuron(s). \textit{Saliency
analysis} identifies parts of the input that
have the largest effect on the output. This consists of a number of approaches that propagate gradients through the
model to identify areas of highest activation and highest sensitivity~\cite{simonyan13,zeiler14, Mahendran16,Selvaraju17,
Sundararajan17}. \textbf{Challenge:} Even simple DL models consist of hundreds of thousands of neurons (e.g. \tabref{tab:data}).
Finding the appropriate set of neurons to visualize can be beyond the powers of human cognition. Saliency analysis works on a per-input-data-item basis; ML practitioners would need specific input data points,
such as images, for these methods. DL data sets consist of tens of thousands of instances,
picking appropriate data instances from these large data sets is imprecise, especially if the data set is new, large and
contains unexplored scientific data.

\subsubsection{Statistical analysis} Many data sets are annotated. Language models are annotated with parts
of speech or linguistic features and image data sets are annotated with object information. For instance, Broden data set~\cite{netdissect17},
has pixel-level annotations that indicate the object to which each pixel belongs. These annotations are used to pose hypotheses
and conduct statistical analyses between neuron(s) activations and annotation to evaluate these
hypotheses~\cite{Sellam19}.
\textbf{Challenge: } Statistical analyses require such annotations to formulate hypotheses. The two data sets we utilize
do not have any annotations. Indeed, most scientific image data sets do not, which makes statistical
analysis impossible.

\section{Workload Characterization}
\seclabel{workload}

We now develop a summary workload that captures the requirements
of a large set of DL model diagnosis queries. Model diagnosis techniques, such as visualization and examination of learned representation,
bring the number of neurons and data instances to be examined to a smaller and manageable number. This section focuses
on queries from these two categories, as these queries helps make downstream analysis, such as feature visualization, attribution and saliency analysis tractable over massive data sets.
We do not include queries from statistical analysis as it requires annotations on the data set.

\begin{table*}[t]
\setlength\tabcolsep{3pt}
\centering
  \begin{tabular}{ p{0.5 cm} p{14 cm} }
  \toprule
  \textbf{QN.} &\textbf{Queries}\\
  \midrule
        Q1. & What is the average value for all neurons for layer $Conv2$ in $model_{A}$ across all classes?~\cite{kahng18,girshick14,raghu17}  \\
        Q2. & What are the top k maximally activated neurons for layer $Conv2$ for all incorrectly classified objects for $model_{A}$?~\cite{kahng18,Liu17,karpathy15}\\
        Q3. & What is the average neuron activation pattern for the last hidden layer in $model_{A}$ for incorrectly classified $class_{a}$ vs.
        correctly classified $class_{a}$?~\cite{kahng18,karpathy15,smilkov17}\\
        Q4. & Compute the similarity between the logits of $class_{a}$ and the representation learned by the last convolution layer by $model_{A}$?~\cite{raghu17,kornbith19,} \\
        Q5. & For images of $class_{a}$ classified as $class_{b}$, what are all of the maximally activated neurons in the last convolutional layer?~\cite{Liu17, kahng18} \\
        Q6. & Does $model_{C}$ learn a representation for $class_{e}$ faster than it learns the representation for $class_{f}$?~\cite{kornblith19, raghu17} \\
        Q7. & How similar are the representations learnt by two different model architectures, $model_{A}$ and $model_{B}$, on the same data set?~\cite{raghu17,kornbith19,Morcos18} \\
  \bottomrule
  \end{tabular}
  \caption{Representative queries for deep learning diagnosis workload.}
\tablabel{tab:sampleQueries}
\vspace{-1.5em}
\end{table*}

An ML practitioner typically starts model diagnosis with techniques utilized by visualization tools from \secref{bground}. They
create data subsets that are incorrectly classified, generating aggregates (such as average activations, top-k highest
activations etc.), and compare them to similar aggregates for data instances that were correctly classified for each class.
They start the analysis from sets~\cite{kahng18,Liu17}, such all incorrectly labeled instances of $class_{a}$, rather than specific instances
to find such patterns. This analysis lets
them identify important patterns for the various subsets and reduce to a manageable number both data instances and parts of the model (layers and neurons) to be examined~\cite{kahng18,Liu17}.
They then start correlating input data and parts of the model,
conducting attribution and saliency analyses.
Similarly, ML practitioners comparing two different models trained on the same data leverage techniques listed in
examining learned representation from \secref{bground}. For instance, they generate neuron activation
values for each data item for both models for each layer. They compare these to the logits for each class learned by the
respective model to decipher each model's rate of learning to understand the impact of additional layers and
their sizes and thus diagnose how complex the model must be to complete this task.


\tabref{tab:sampleQueries} lists representative queries from the literature used to diagnose DL models. We make two observations
from this list of queries. First, DL model diagnosis queries requrie one of three quantities: the top-k maximally activated neurons, the distribution of maximally activated neurons or the average activation values. The focus on maximal and top-k values as opposed to minimal values is due to activation functions~\cite{afunc}
used in DL models. Without such functions DL models are just complicated linear regression models. ReLU is the most commonly used activation function today~\cite{relu}.
It removes negative values and propagates positive values. Mathematically, ReLU is defined as $max(0,val)$. Therefore, in the DL
literature sample queries often focus on average or maximal values but not minimum values. Thus, to characterize an ML diagnosis workload instead of focusing on all aggregates we focus on three aggregates
(1)Top-k maximally activated neurons, (2) Average activation values for neurons  and (3) distribution of maximally activated
neurons.

Second, each query in \tabref{tab:sampleQueries} is part of a family of queries. For instance, the answer
to Q1 requires average values of all neurons for a specific layer ($conv2$) for all classes. A family of queries for Q1
would include average values of neurons for \textit{any layer} and \textit{any class} where data instances could be \textit{correctly} or \textit{incorrectly} classified.
We can see that queries Q3 and Q1 belong to the same family. Similarly, Q2 belongs to a family of queries that require top-N neurons, across classes, layers, incorrect, and correct classification.
Thus, to charecterize this workload we utilize the entire family of queries. We call
these families of queries \textit{query sets}.

We now introduce some notation and define query sets formally.

A DL model $M$ is a vector of $N$ units or neurons.
$M$ is learned and tested over data $D$. Artifacts, such as activations $A$, are vectors of the
same dimensionality as $M$, computed over data $D$. $a_{id}$ are the activation value(s) of $i$  neuron(s), where $i \subseteq N$,
on $d$ data item(s), where $d \subseteq D$. A query set $S$ computes a measure $\phi$, such as the  mean, top-K, count, or count of maximum values for $a_{id}$ etc.
Given the preceding notation we can define a DL model diagnosis query set:

\begin{mydef}
    A query set $S(a_{id},\phi)$ is a set of queries, where $i \subseteq N, d \subseteq D$, and $\phi$ is a measure.
\end{mydef}

Instead of evaluating our techniques on specific queries from \tabref{tab:sampleQueries}, we utilize the three query sets shown in \tabref{tab:summaries}
to characterize DL model diagnosis workload. These query sets include all queries of a specific family. We leverage these query sets to measure effectiveness of
sampling techniques to ensure these techniques do well on the entire family of queries represented by the query set, not just on individual queries.

\begin{table}[h]
    \centering
    \begin{tabular}{ p{7 cm} }
    \toprule
    \textbf{Query Sets}\\
     \midrule
    \textbf{S1.} Set of top-K maximally activated neurons. \\
    \textbf{S2.} Average activation values of neurons. \\
    \textbf{S3.} Distribution of maximally activated neurons. \\
    \bottomrule
    \end{tabular}
        \caption{Query sets for deep learning model diagnosis workload. }
    \tablabel{tab:summaries}
    \vspace{-2.0em}
\end{table}

Query Q2 and all queries of this family are represented by query set S1, which computes the top-K maximally activated neurons. Queries Q1, Q3, and others of this family are represented by query set S2, which computes the average activation for neurons. Queries Q4, Q5, Q6 and Q7, and others of their family are jointly represented by query sets S2 and S3, because finding similarity
depends on the average neuron values and the maximally activated neuron distribution.

Query sets can consist of any combination of neurons and data items.
Instead of considering this immense set of combinations, we limit our evaluation to all combinations of layer, class and classification (correct or incorrect).
Thus, to measure accuracy of a query set for a sample, we first compute the query results for each of these combinations (layer, class and classification).
Next, we compute a metric comparing the results from the sample with the results for the same combination on the entire data set.
Our comparison utilizes metrics specific to each query set, e.g., precision for S1,
cosine distance for S2 and, Jensen-Shannon distance for S3 (we describe these metrics and the rationale for picking them in more detail in \secref{eval}).
Finally, we calculate the over-all query set accuracy for each query set by averaging the value of the corresponding metric over the combinations.

\section{Approach}
\seclabel{sys}

To enable interactive model diagnosis, our approach creates a sample. We compute the results of a query set on
this sample instead of entire data. In this section we describe our approach and present other baseline techniques for selecting
these samples.

The key insight we utilize to avoid generating and storing activation values is that DL models learn a lower
dimension representation of the data, and a classifier.  DL training transforms the input data, creating a new representation with
each layer. Training criteria encourage training set neighbors, such as data points from the same class, to have similar representations.
Leveraging this lower dimensional representation learned by the model
has the dual benefit of reducing dimensionality of the data and focusing on the representation learned by the model.
Since the objective of the workload is to diagnose this model, we hypothesize that leveraging the learned latent space
to select a sample will be key to understanding
what the model has learned. For model diagnosis we view the training, and test data points in the
 \textbf{latent space}, i.e., instead of viewing the data in the high
 dimensional original format of images, audio or text, we utilize this lower dimensional representation of the data learned
 by the model's last hidden layer to create samples.

Our goal is to diagnose the model, which implies that a subset of the queries will focus on what the model
got wrong, as seen in \tabref{tab:sampleQueries}. In a classification problem with multiple classes, the decision boundary
partitions the underlying vector space into multiple regions, one for each class.
Decision boundaries are where the output label of a classifier is ambiguous, i.e., where
errors and mis-classifications occur. The diagnosis of a DL model requires exploration of the \textbf{decision boundary} for a model~\cite{Guidotti18,Wu18}.

\begin{figure*}[!tbp]
  \centering
  \includegraphics[width=0.49\linewidth]{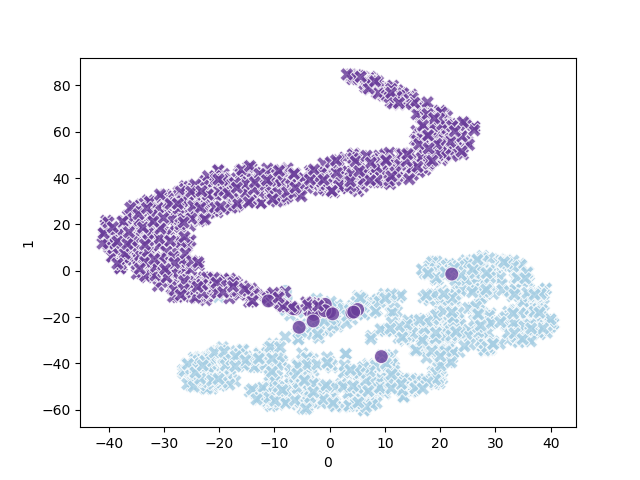}
  \includegraphics[width=0.49\linewidth]{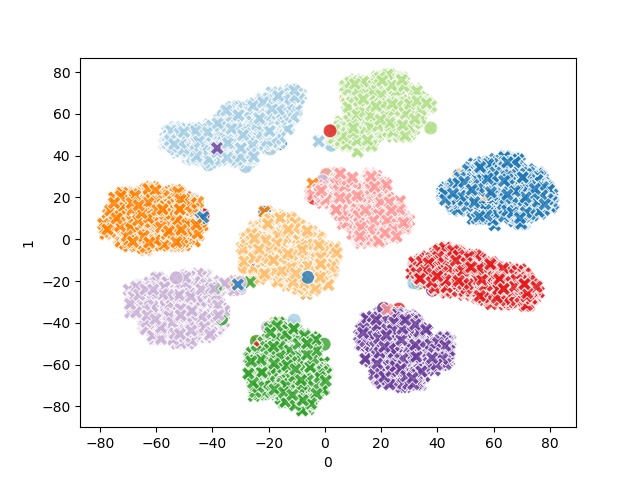}
  \caption{T-SNE representation of test data or Galaxy Zoo2 (left) and MNIST (right)
  from the last hidden layer. Each data point represents an input
  image from the test set. Data point colors represent
  the true labels.} \figlabel{fig:tnseAll}
\end{figure*}

For instance, \figref{fig:tnseAll} depicts this lower dimensional representation for two data sets we use for evaluation, MNIST~\cite{mnist} and Galaxy
 Zoo2~\cite{gzoo}. We use a dimensionality reduction technique, t-Distributed
Stochastic Neighbor Embedding (t-SNE)~\cite{tsne}, to reduce dimensions of this data to visualize it in two dimensions.
In \figref{fig:tnseAll}, each point represents an image from the test set, and colors indicate the true class labels.
We make two observations from this visualization:
(1) the data representations in latent space show separation for each class, and
(2) most mis-classified instances are on the edges of data points groupings.

For DL model diagnosis, the top-k maximally activated neurons and distribution of maximally activated neurons
form a large subset of queries in addition to the average values of neurons.
The top-k queries are an important area of database research. The best known general-purpose algorithm for identifying top-k items is
the Threshold Algorithm~\cite{fagin03}, which operates on sorted multi-dimensional data required
to compute the top-k elements. Approximate algorithms for top-k retrieval require building probabilistic models to fit the score
distribution of the underlying data as proposed in ~\cite{theobald04}. However, we wish to avoid computing and
storing activations for the entire data set.

Thus, our approach is based on utilizing the lower dimensional representation when selecting data items for
our sample, and focusing on decision boundaries in the \textit{latent space} when selecting the data points to include in our sample.

\subsection{Baselines}
\seclabel{baseline}
The naive way of selecting a sample that covers the n-dimensional latent space is to create a grid in that space and sample from each
each partition. We sample here from the latent space; our goal is to ensure our sample contains instances of data that lay in different regions of that latent space.
However, the latent space we choose is high dimensional, e.g., for the MNIST dataset, the latent space is $84 D$. Even if we divide each dimension into two buckets, we get a total of $2^{84} \sim 1.93e+25$ buckets. Instead, we reduce the dimensionality of data in the latent space for this analysis using PCA. We call this naive technique
\textit{simple latent space} sampling. For this sampling technique, we collect equal number of instances
at random from each underlying grid.

Another way to lower the dimensions is to utilize the classification result.
Each data point is classified by the model as belonging to a class. This result is encapsulated in a
\textit{confusion matrix} (a.k.a. the error matrix), which tabulates the performance
of a classification algorithm. For a binary classifier, the
confusion matrix counts the number of true positives, false positives,
true negatives, and false negatives. For multiple labels,
the confusion matrix generalizes this concept. Each row of the matrix
represents a predicted class, while each column represents a
true class. In this technique, we sample based on cells in the confusion matrix.
We call this technique \textit{stratified by confusion matrix (CM)}.

In database systems such as BlinkDB\cite{Agarwal13}, strata are
defined over a subset of columns that typically correspond to categorical
valued attributes, e.g. \texttt{city}.  For DL model diagnosis, the
underlying data can be considered a relation, with each row
representing a data item (e.g., one image) and each column
a value of interest, such as the activation value for a neuron in the
model.  Each row can be extended with metadata, such as
the predicted class and the class label. The
\textit{stratified by CM} sample thus serves as a stratified sample baseline.

In addition, we use two other techniques from the literature as baselines.
First, we use  visualization aware sampling (VAS) for large scale data visualization, such as scatter
and map-plots. VAS is based on the interchange algorithm~\cite{Park16}, which
selects tuples that minimize a visualization-inspired loss function. Visualization-inspired loss is
based on three common visualization goals: regression, density estimation and clustering. The interchange algorithm
creates a sample that maximixes visual fidelity of the data at arbitrary zoom levels.

Second, we use explicable boundary (EB) trees~\cite{Wu2017} to create a single sample from input
data.  This method constructs a boundary tree to approximate the complicated deep
neural network models with high fidelity. EB trees provide a single sample for a dataset and a model which explains the
boundary between each class learned by the DL model.

\subsection{Clustering in Latent Space}
An important part of our approach to selecting a sample for DL model diagnosis is to
ensure that model decision boundaries are represented in the sample. To determine boundaries
in latent space, we cluster data in latent space and fit a model to estimate the parameters for each class
in that space. We do this in both supervised and unsupervised manner. When fitting
a supervised model, we use the class labels. In the unsupervised case, we use parameterized models
so we utilize the number of unique classes present.

In both supervised and unsupervised cases the models fitted to the latent space provide us
with the likelihood that and object belongs to a class or cluster.
For binary classification to determine whether an object belongs to class $A$ or class $B$,
let $P(A|x_{i})$ be the likelihood that a data instance $x_{i}$ belongs to class $A$. In this case, the points on the decision
boundary of class A and class B are those for which the ratio $\frac{P(A|x_{i})}{P(B|x_{i})}$ is $\approx 1$. A lower
value of likelihood ratio would imply that $P(B|x_{i}) > P(A|x_{i})$ in which case $x_{i}$ would be assigned to cluster or class $B$.
The higher the likelihood that an object belongs to class $A$, the higher the ratio $\frac{P(A|x{i})}{P(B|x_{i})}$ will be.

For a multi-class classifier, where a data point $x_{i}$ may belong
to classes $\subset a,b,c,\dots $, this ratio  would be,
$\frac{P(A|x_{i})}{\sum_{z\subset b,c,d, ...}P(Z|x_{i})}$, or
$$\frac{P(A| x_{i})}{P(\neg A | x_{i})}$$

Our sampling technique clusters the data in the latent space, then sorts data in
each cluster or class by the ratio of likelihood of belonging to that
particular class. This sorted list thus consists of exemplars on the higher end
and outliers on the lower end of the list. We utilize a tuning parameter $j$ to
determine the proportion of exemplars and outliers in our sample. We select $j\%$ from the outliers
and $1-j\%$ from the exemplars. Algorithm~\ref{alg:sampl} describes this approach in further detail.

\begin{algorithm}
\DontPrintSemicolon
\SetAlgoLined
\SetKwInOut{Data}{Data}
\SetKwProg{Def}{def}{:}{}
\Data{input data in latent space,  $f$,$k$, $j$}
\tcp{k num class labels, $f$ is sample size}
$Clusters\leftarrow None$    \;
$sample\leftarrow None$    \;
Clusters = ClusterAndSortData(data,k)\;
\ForEach{$cluster_{i}$ in Clusters}{
    $s1 \leftarrow  data.head(f * j)$ \;
    $s2 \leftarrow  data.tail(f * ( 1 - j ))$ \;
        $sample \leftarrow s1 + s2 + sample$\;
}
return sample\;
\BlankLine
\caption{Clustering }\label{alg:sampl}
\end{algorithm}

For the unsupervised technique, we utilize a parameterized clustering technique, the Gaussian Mixture Model (GMM).
These models offer a probabilistic way to represent normally distributed sub-populations within an overall
population. We set the number of clusters in GMM to be equal to the number of unique classes in the dataset.
We utilize variational estimation for the GMM ~\cite{attias00}, where the effective number of
components can be inferred from the data.

For the supervised technique, we use max-margin classifiers to classify the data in the latent space. Margin classifiers
are a class of supervised classification algorithms that utilize distance from the decision boundary
to bound the classifier's generalization of error. Support vector machine (SVM)~\cite{svm}
is an example of this category of classifiers, which learns boundaries based on labels
so that the examples of the separate classes are divided by a clear
gap that is as wide as possible. SVMs utilize kernel functions~\cite{kernel}; these help to projecting data
to a higher dimensional space where points can be linearly separated. DL models
do not have non-linear activation functions after the last hidden layer, so the latent representation
from last the hidden layer should enable discovery of linear boundaries. Thus, we utilize a linear
kernel for SVM~\cite{liblinear}, which has the dual advantage of being faster than non-linear kernels
and less prone to over-fitting. Results of the classifier are turned into a probability distribution over classes by using
Platt scaling~\cite{platt99,wu04}. These probabilities are used to sort the data items in each cluster or predicted class
and then select a sample.

\section{Evaluation}
\seclabel{eval}

Here, we empirically evaluate our hypotheses from our sampling
approach, namely sampling evenly from the latent space is not sufficient; model decision boundaries are the most important
region of this latent space for answering
model diagnosis queries; and they must be well represented in a reliable sample.

We evaluated our sampling techniques from \secref{sys} on two different
data sets. We first describe metrics we used to evaluate query sets defined in \secref{workload} and
data sets and DL models we used for experimental evaluation. We then describe the experiments we conducted and
analyze their results.

\subsection{Metrics}

\textbf{Query set S1} retrieves the set of top-K maximally activated neurons. To measure how well our sample performs we use
\textit{precision} as the metric. Precision is the fraction of top-k results from the sample
that belong to the true top-k result. Precision lies between $[0,1]$. A precision value of 0 implies that the sample top-k does not
contain any of the full data top-k neurons.

\textbf{Query set S2} retrieves the average value of neurons. This is a high dimension vector of floating points,
where dimension is the number of neurons in a layer. Additionally, this is a sparse vector, i.e., many neurons may
have zero average activation because of non-linear activation like ReLU. We used cosine distance~\cite{cosinedist} to
measure the distance between the average vector for the entire dataset and the average vector from sample due to
the properties of high dimensionality and sparseness of the average neuron vectors, which lies between $[0,1]$.
Cosine distance is defined as:
$$1 - \frac{A.B}{\|A\| \|B\|}$$

\textbf{Query set S3} retrieves the distribution of maximally activated neurons. As this is a true
distribution an obvious metric would be Kullback-Leibler (KL) divergence~\cite{kldiv}. However, we encounter
two issues with using this metric. First, KL divergence is unbounded, which means it is not a true
metric, and it is difficult to assess how close two distributions were. Second, KL divergence is defined only on distributions with non-zero entries. This is
not true for maximally activated neuron distribution, which may have neurons with zero counts.
Thus, we used Jensen-Shannon divergence~\cite{jsdist} instead, which is based on KL divergence. Jensen-shannon divergence is
both symmetric and finite valued. Jensen-Shannon distance is squareroot of Jensen-Shannon divergence which is defined
as:
$$\sqrt{\frac{D(p\|m)+D(q\|m)}{2}}$$
and lies in $[0,1]$ .

\subsection{Datasets and Models}
We evaluated our sampling techniques on
two data sets, Galaxy Zoo2~\cite{gzoo} and MNIST~\cite{mnist}. For each data set, we built and evaluated one
deep learning model.
The MNIST data set consists of $28x28$ pixel
gray-scale images of handwritten numerical digits with a training and test set
of 60K and 10K images, respectively. We trained the six
layer neural network depicted in \figref{fig:mnist_model}. This model
is a based on LeNet-5~\cite{Lecun98} for classifying MNIST data set with added batch-normalization after every layer.

Galaxy Zoo2~\cite{gzoo} is a public catalog of
$\sim 240,000$ galaxies from the Sloan Digital Sky Survey~\cite{sdss} with classifications
from citizen scientists. The Galaxy Zoo decision tree~\cite{desctree} lists the questions answered by citizen
scientists. We took a subset of this data set to classify images that appear edge-on vs face-on (question T01 in~\cite{desctree}).
The training and test data sets consist of $54,333$ and $2118$ images, respectively,
each a $69x69$ color image. We trained a model depicted in \figref{fig:edge_model}, which is a
variation of the model from~\cite{Dominguez18}. In our variation of this model, we reduced the number of dropout layers
and added batch normalization after every convolutional layer. We achieved $99\%$ accuracy on
the test set and an overall weighted F1 score of $0.99$.

\begin{figure}[t]
        \includegraphics[width=0.95\columnwidth]{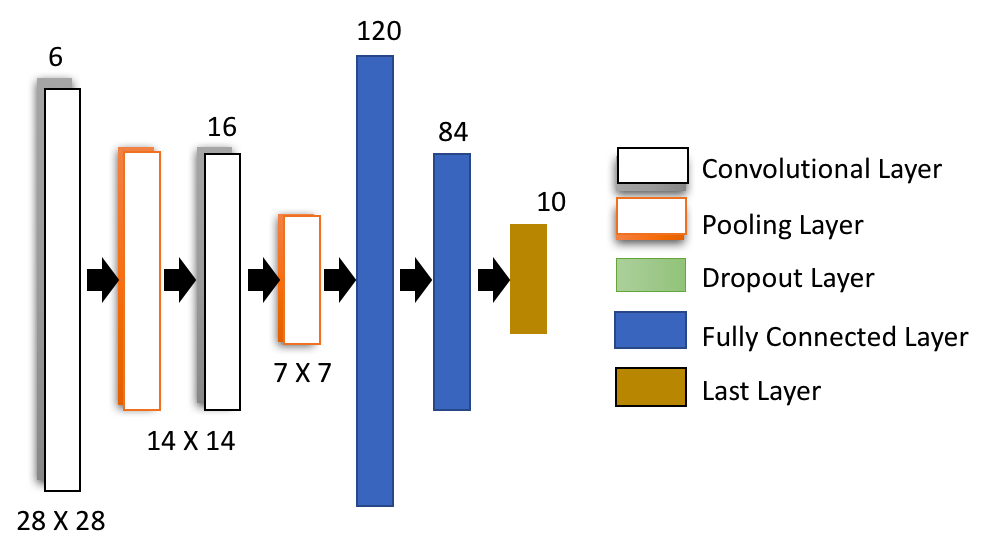}
\caption{Deep learning model for MNIST data set.}
    \figlabel{fig:mnist_model}
\end{figure}

\begin{figure}[t]
        \includegraphics[width=0.95\columnwidth]{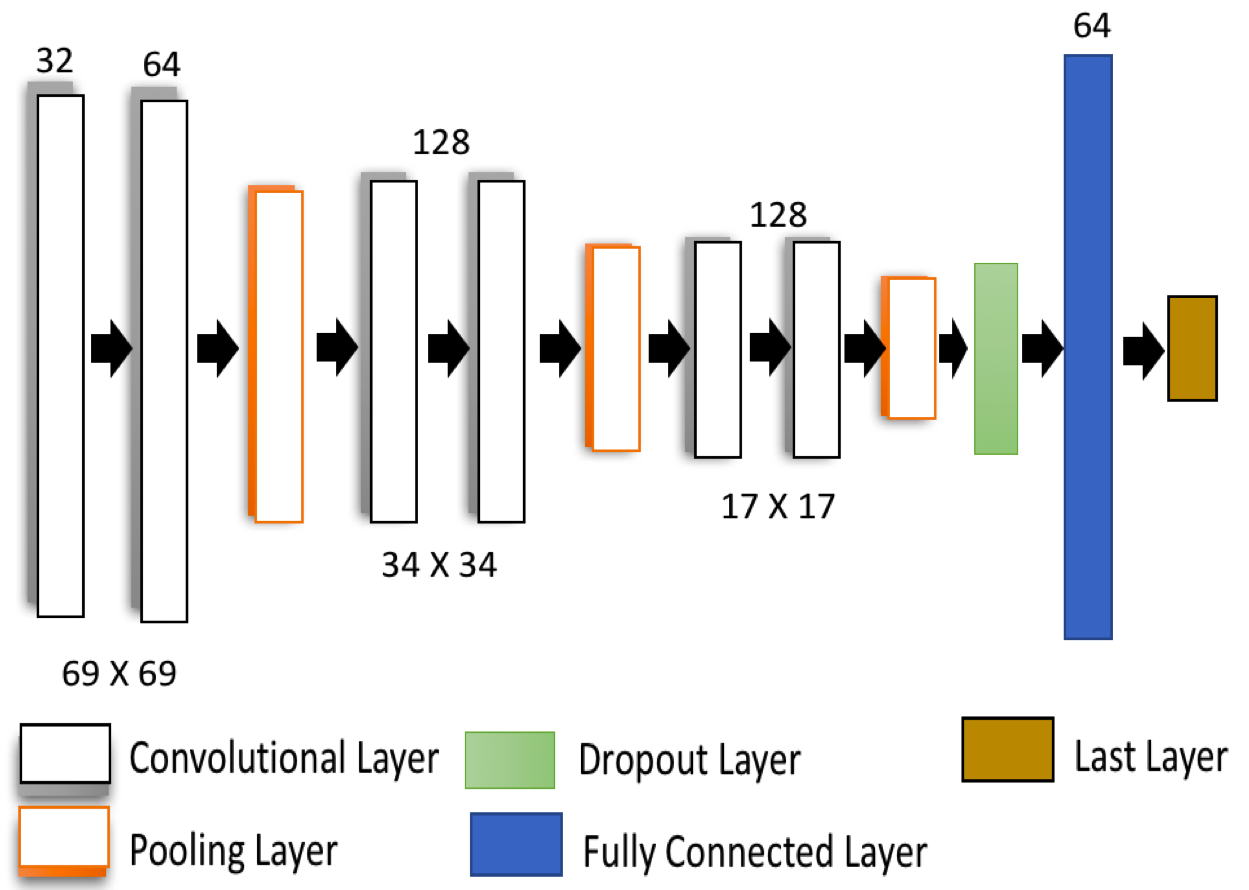}
\caption{Deep learning model for galaxy Zoo2 data set.}
    \figlabel{fig:edge_model}
\end{figure}

We trained the models and extract activations from them using the
TensorFlow~\cite{tensorflow} library. For both models, we used the representation from the last hidden
layer to drive our sampling technique, and
the last hidden layer was a fully connected (FC) layer. The MNIST data representation
is from layer FC-2 (\figref{fig:mnist_model}) with 84 neurons. The
Galaxy Zoo2 data representation is from layer FC-1
(\figref{fig:edge_model}) with 64 neurons.

\subsection{Experiments}

\begin{figure*}[t]
    \begin{subfigure}{\linewidth}
    \begin{tabular}{cccc}
        \includegraphics[width=0.33\linewidth]{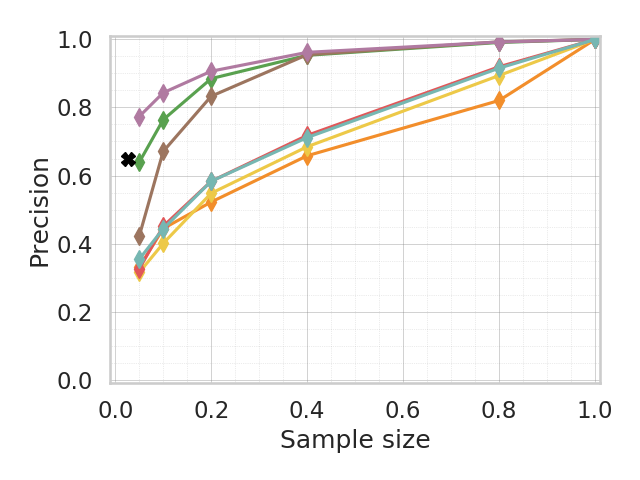}
        \includegraphics[width=0.33\linewidth]{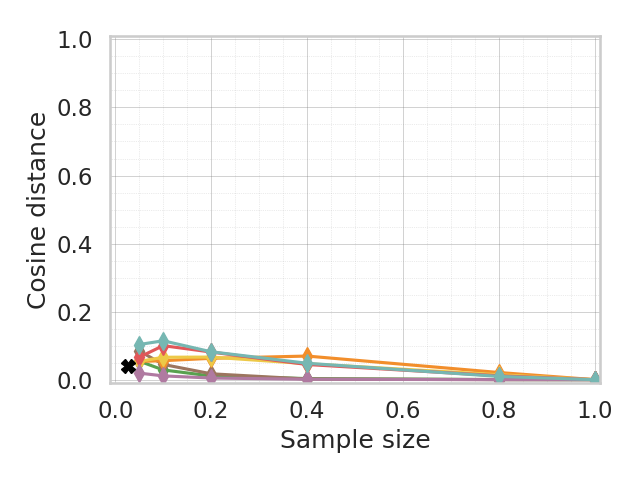}
        \includegraphics[width=0.33\linewidth]{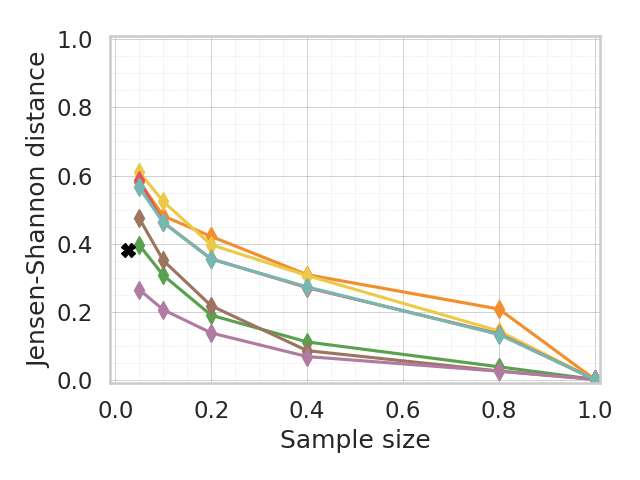}\\
        \includegraphics[width=0.33\linewidth]{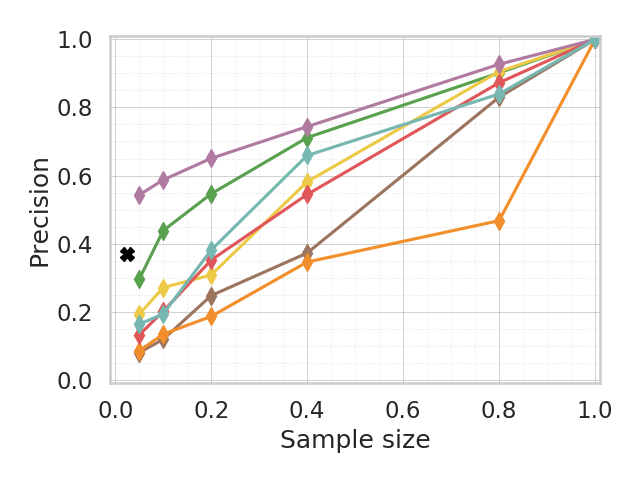}
        \includegraphics[width=0.33\linewidth]{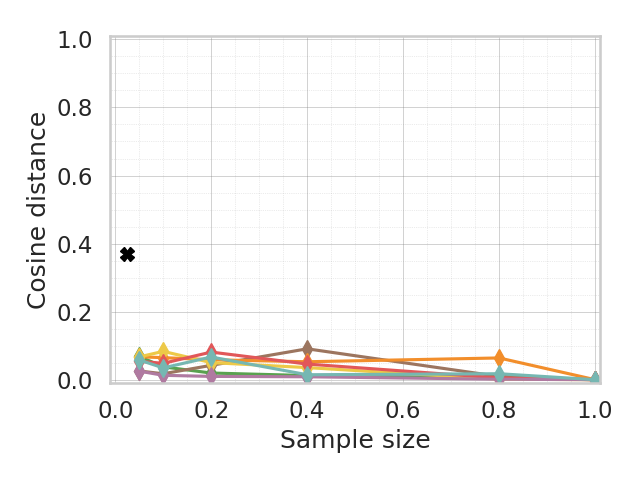}
        \includegraphics[width=0.33\linewidth]{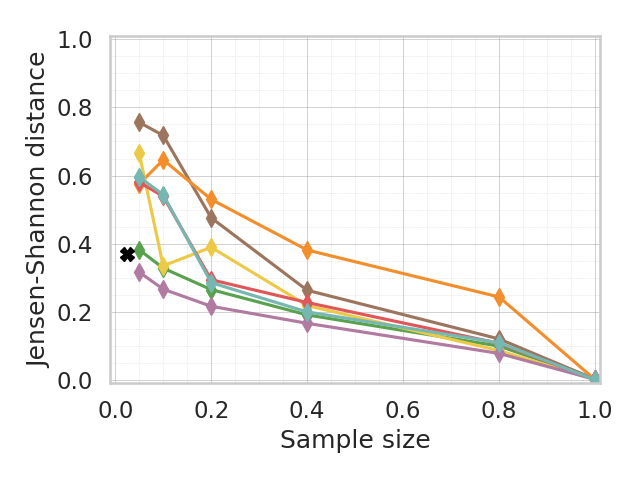}\\
    \end{tabular}
    \end{subfigure}
    \begin{subfigure}{\linewidth}
        \centering
        \includegraphics[width=0.78\linewidth]{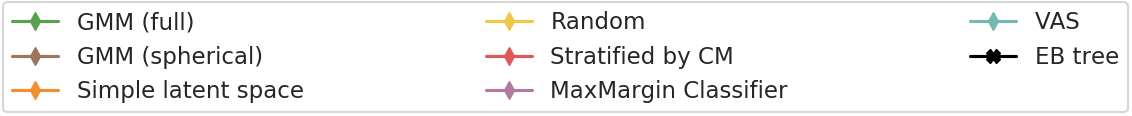}
    \end{subfigure}
    \scriptsize
    \caption{ Metrics for S1, S2 and S3 for increasing sample
    size for various sampling strategies. Top row MNIST, bottom row Galazy Zoo2. From left to right columns, S1, S2, and S3.
    The X-axis shows the sample size as a fraction of the entire data set.}
    \figlabel{fig:ss_res}
\end{figure*}

For our first experiment, we evaluated the three query sets on the two data sets
using the metrics described above. For each data set, we created samples of size 5\%, 10\%, 20\%, 40\% and 80\%
for the eight sampling techniques we are evaluating. Our rationale for choosing sampling techniques is described in \secref{sys}, here, we provide a brief description of each techniques:

(1) \textit{Random} sampling draws a sample from the data set uniformly at random without replacement.
(2) \textit{Stratified by CM} sampling contains a sample with data items drawn from each cell of the confusion matrix in proportion
to the  number of data items in the cell.
(3 and 4) \textit{Visually aware sampling (VAS)}  and \textit{Explicable Boundary (EB) tree} sampling utilize the techniques
specified in~\cite{Park16} and~\cite{Wu18} respectively.
(5) \textit{Simple latent space} sampling divides the latent space into
a grid and then samples equally from each cell in the grid.
(6 and 7) \textit{GMM} sampling fits a GMM to the data points in latent space. For
each of the resulting clusters, data points belonging to each cluster are sorted
by the likelihood ratio $\frac{P(A|x)}{P(\neg A| x)}$ of belonging to that cluster. The sample
is then created by selecting data points from the two ends of this list for each cluster, with a tuning factor $j$ determining what fraction is selected from either end.
We have two GMM samples since we evaluated impact of two types of co-variance matrix, spherical and full.
Finally, (8) \textit{MaxMargin classification} sampling classifies the data points in the latent space
with a max margin classifier, sorting points in each class by the ratio of their likelihood belonging to that
class, and choosing from the two ends of this list with a tuning factor of $j$, like the GMM samples.

For the first experiment GMM  and MaxMargin samples -  we fixed the tuning factor $j$ to 0.7. We studied the
impact of this tuning factor in the third experiment. The EB tree technique creates a single sample
since there is a single boundary tree for a model and corresponding data. \figref{fig:ss_res} shows the results of this experiment for both data sets.
As we increased sample size the query set results got increasingly more accurate until,
at fraction $1.0$ or on the full data set, the metrics for all sampling techniques were coincident at 1.0 for S1 and 0 for S2 and S3.

For \textit{simple latent space} sampling we reduced the dimensionality of latent space from 84 and 64 to 5 for both
MNIST and Galaxy Zoo2 and then divided each dimension into 2 bins, resulting in $2^5$ or $32$ bins.
We then sampled equally from each bin. This is the only technique where we sampled equally
rather than sample in proportion to the number of items in the bin. We did this in order to
evaluate the impact of sampling from the latent space. Interestingly, this technique did not do well on all three
query sets. To minimize the impact of randomness, we selected each sample ten times and evaluated it and average
results from these ten iterations. As we can see from \figref{fig:ss_res}, the \textit{simple latent space} sample behaved as well as the random sample.
While this sample provided adequate results on S2, giving on average less than 10\% error, its performance on
S1 and S3 was not adequate. The knee seen for this sample (at 80\% of the data set) occurred because at this point the sample had the fewest
number of data items compared to other samples: data were unevenly distributed in the latent space, and
we sampled equally from each bin rather than in proportion to the size of the bin.

The \textit{stratified by CM} sample performed much better than both the \textit{random} and \textit{simple latent space}
samples for S1 and S3. This is because accuracy over query set is determined by averaging query results from each layer, for each class, and for correctly and incorrectly
classified data items. If a sample did not contain any instances of incorrectly classified data items from $class_{a}$, for instance, then the query result was set to 0 for S1 and 1 for S2 and S3.
As \tabref{tab:sampleNumbers} shows, \textit{random} and \textit{simple latent space} samples often do not have any data items from the
incorrectly classified data items. \textit{VAS} did as well as \textit{stratified by CM}, this is of note because the VAS sample had no knowledge of classification
of each data points and was trying to minimize a visualization-based loss function, which is trying to ensure that
that the sample replicated the data density of the original distribution.

All three clustering-based samples \textit{GMM (full)}, \textit{GMM (spherical)} and \textit{MaxMargin classification} based samples did better than the baseline samples on all three query sets in most cases. \textit{GMM (full)} did better than \textit{GMM (spherical)} for both data sets.
\textit{GMM (full)} fit the data better, as expected, and thus did better on selecting exemplars and outliers when compared to \textit{GMM (spherical)}.
The goodness of fit is dependent on the data set. \textit{GMM (spherical)} does better than \textit{stratified by CM} for the MNIST
data set but worse for the Galaxy Zoo2 data set. From the two dimensional representation of the data in latent space for
the two data sets in \figref{fig:tnseAll} we can see a separation between the ten clusters in MNIST, while the two clusters in the Galaxy Zoo2
data set were not clearly separated. Additionally, for MNIST each cluster appeared to be somewhat symmetrical, but
the two clusters for the Galaxy Zoo2 data set did not have a clear separation, and one of the clusters is highly asymmetrical.
\textit{GMM (spherical)} with a isotropic co-variance matrix has difficulty fitting the Galaxy Zoo2 data set.
\textit{GMM (full)} fit a more complex gaussian to each cluster, and this, in turn, provided a much better
estimation of outliers vs exemplars. This difference can be seen in two data sets. While \textit{GMM (full)} sample
did better than \textit{GMM (spherical)} sample for both data sets, the difference in performance was higher
for when the underlying data distribution assumptions were not met for \textit{GMM (spherical)}.
\textit{MaxMargin classifier}-based sampling performed the best on all three query sets. This implies that this
technique could distinguish between exemplars and outliers better than GMM, and a sample based on the \textit{MaxMargin classifier}
is better suited to addressing queries for DL model diagnosis. This is further supported by examination of the data items
selected by each sample. \tabref{tab:sampleNumbers} shows the number of correctly and incorrectly classified data items
selected by each sampling technique on a 5\% sample. \textit{MaxMargin classification}-based sampling
selected the highest number of mis-classified data instances. These results for S1 and S3 support our hypothesis that emphasis on the decision  boundary improves samples
for all query sets.
Finally, \textit{EB tree} technique provided a single sample since there was only one boundary for model. As expected,
it did well picking the outliers and therefore performed well on both S1 and S3. For both data sets, EB-tree based sample
was the smallest and performed second best on these two query sets. However, as the EB tree sample focused inordinately
on the outliers, it did not perform as well on S2. On the well-separated latent space for the MNIST
data set the \textit{EB-tree} performed on par with other sampling techniques. However, for the Galaxy Zoo2 data set, it did not perform as well.
The \textit{MaxMargin classification}-based sampling performs better than EB-tree sample for all three queries for both data sets.

\begin{figure*}[t]
    \centering
    \begin{subfigure}[b]{0.98\linewidth}
        \includegraphics[width=\linewidth]{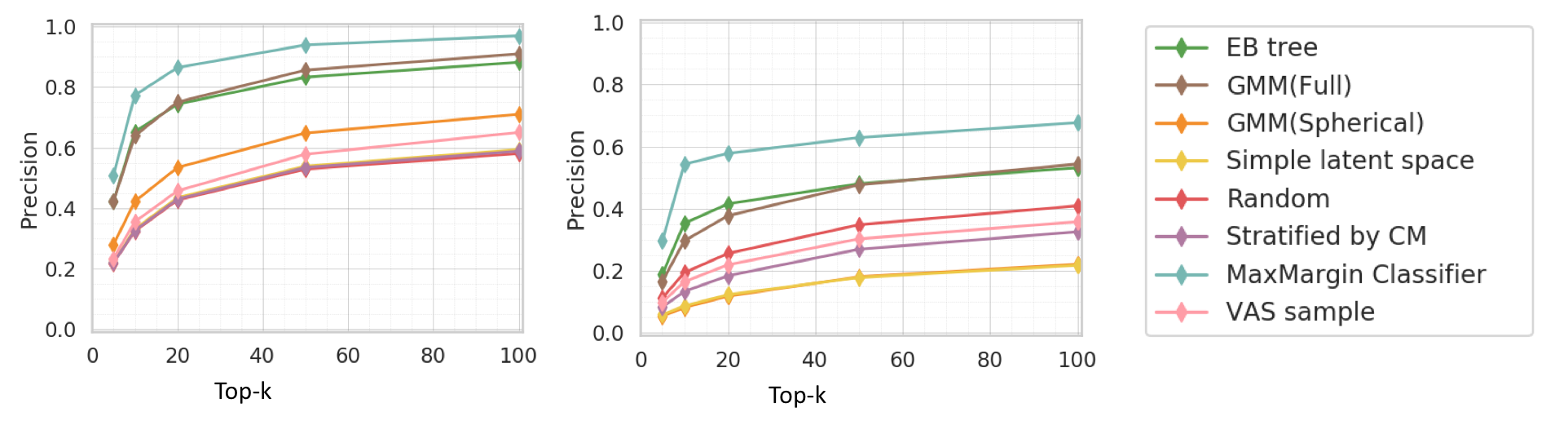}
    \end{subfigure}

    \scriptsize
    \caption{ Metrics for S1, number of top-k neurons in the 5\% sample. Left panel MNIST, right panel Galaxy Zoo2.}
    \figlabel{fig:top_kres}
\end{figure*}

In the second experiment, we examined the impact of varying the number of top-k neurons in S1 and measured the precision
achieved by each of the eight sampling techniques. We varied the nuResults for this experiment are shown in \figref{fig:top_kres}.

For the MNIST dataset, a $5\%$ sample had a precision $0.98$ for the top-100 neurons. However for the Galaxy Zoo2 data set this
number was much lower, at $0.70$ for the top-100 neurons. This is due to two factors:
(1) the test data set, over which we evaluated this query for MNIST was 10k while for the Galaxy Zoo2 data set it is 2k. A 5\% sample
was 500 data items for MNIST and 105 items for Galaxy Zoo2 data set.
(2) the  model for MNIST had 107,786 parameters or neurons, and Galaxy Zoo2 had an order more parameters at 1,095,842.

Thus, a $5\%$ sample for the Galaxy Zoo2 data set was both smaller and trying to capture a more complex model.
This is confirmed by an additional experiment, where we increase the Galaxy Zoo2 sample size to 500 elements, we get 85\% coverage
on the top-100 neurons.

For both data sets \textit{MaxMargin classification} sampling had the highest precision. \textit{EB tree} was next for both data sets. This is
not surprising because EB tree focuses on decision boundaries. This reinforces our hypothesis that decision boundaries
need to be well represented for a sample to perform well on model diagnosis queries.

\begin{figure*}[t]
    \begin{subfigure}{\linewidth}
    \begin{tabular}{cccc}
        \includegraphics[width=0.33\linewidth]{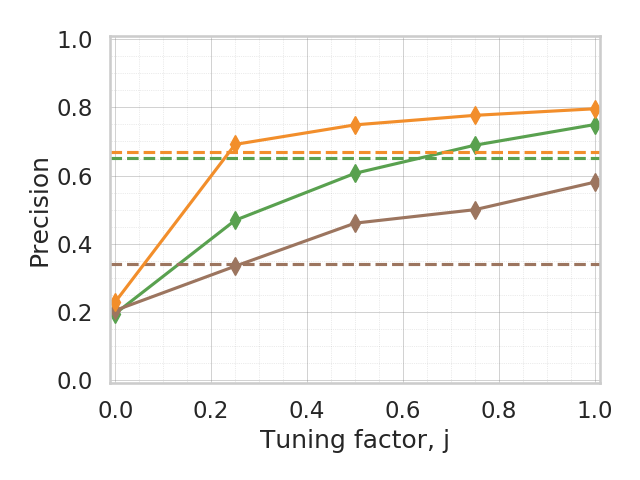}
        \includegraphics[width=0.33\linewidth]{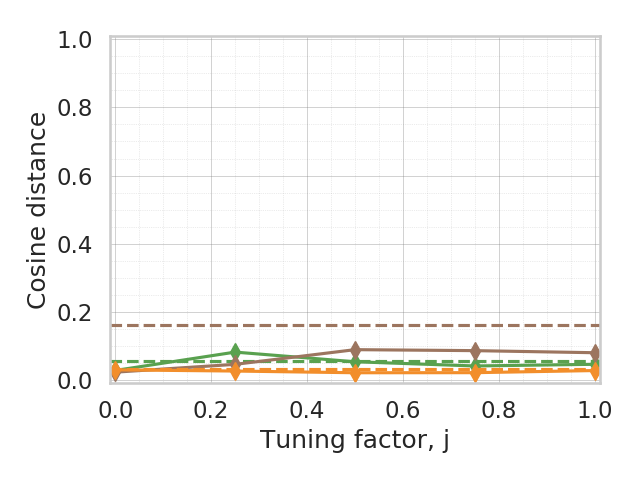}
        \includegraphics[width=0.33\linewidth]{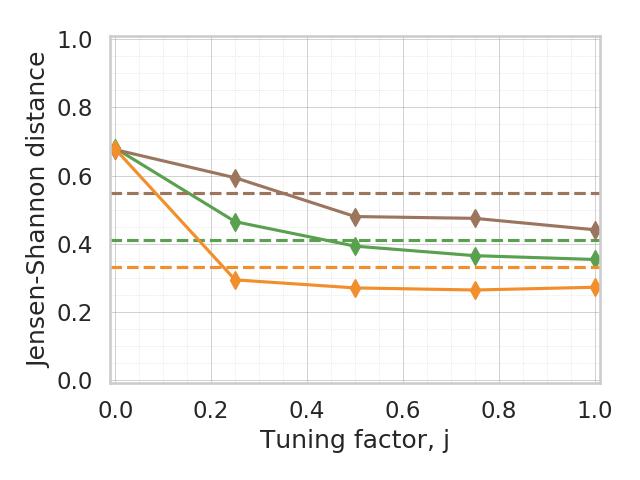}\\
        \includegraphics[width=0.33\linewidth]{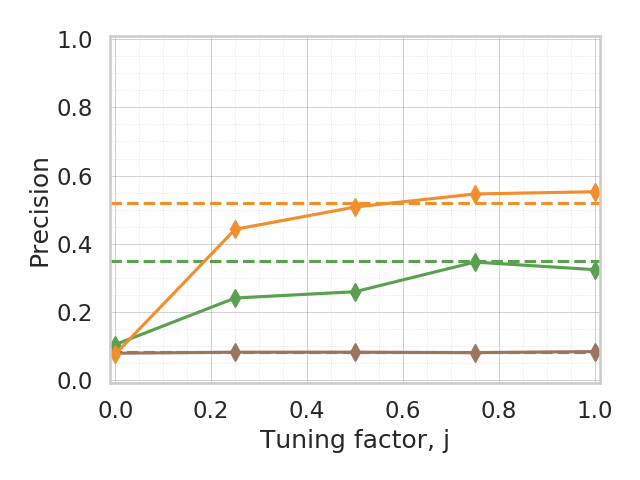}
        \includegraphics[width=0.33\linewidth]{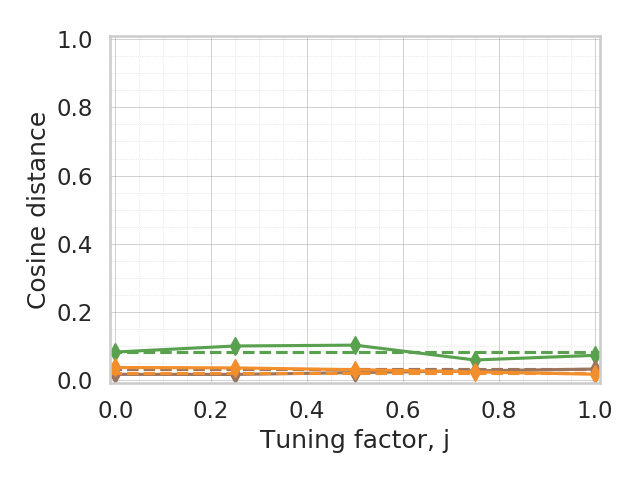}
        \includegraphics[width=0.33\linewidth]{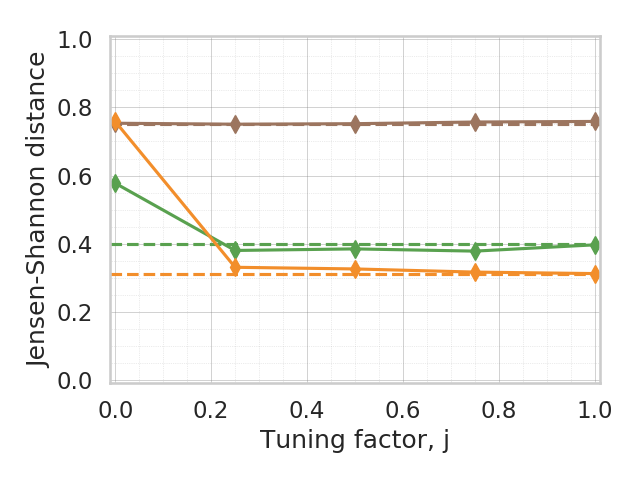}\\
    \end{tabular}
    \end{subfigure}
    \begin{subfigure}{\linewidth}
        \centering
        \includegraphics[width=0.78\linewidth]{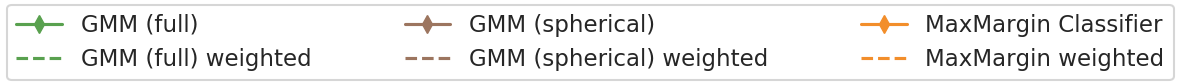}
    \end{subfigure}
    \scriptsize
    \caption{ Impact of tuning factor $j$ (x-axis) on metrics for
    MaxMargin and GMM based sampling strategies. From top row MNIST, bottom row Galazy Zoo2, from left to right columns S1, S2, and S3.}
    \figlabel{fig:tf_res}
\end{figure*}

In the third and final experiment, we evaluated the impact of tuning factor $j$ for three clustering samples \textit{GMM (full), GMM (spherical), MaxMargin}.
Tuning factor $j$ is a number between 0 and 1 and is used to determine how many data points in the samples come from the lowest
values of likelihood ratio or outliers. We evaluated the impact of this tuning factor on a $5\%$ sample for all three sampling strategies.
We evaluate query sets S1, S2 and S3 on tuning factor values of 0, 0.25, 0.50, 0.75 and 1.00.
In all three sampling strategies, we picked data items in order from the sorted list for each cluster.
Our sample is selected by selecting items from both ends of the sorted list and picking $frac * j$ items from the head or outlier
end of the list and, $frac*(1-j)$ from the exemplar end of the list. Thus, for the tuning factor value of 0,
all data instances in the sample are picked from the exemplar end of the list and for
a tuning factor value of 1 all data instances were picked from the outlier end of the list. In this experiment, we additionally
created a weighted sample, where the weight was simply the reciprocal of the likelihood ratio.
 Likelihood ratio can be unbounded for exemplars, therefore for purposes
of numerical stability we selected a threshold. To reduce the impact of random selection, we selected a weighted sample ten times and reported the
average value. \figref{fig:tf_res} shows the results of this experiment.
For S1, when the data set contained only exemplars at tuning factor 0, precision was the lowest for the sample. Precision grew as the value of tuning factor
increased and plateaued at tuning factor $\sim 0.7$.
The max top-10 precision for the MNIST dataset was 0.8 and galaxy Zoo2 is 0.57. This was the  max value for top-10 that can be achieved
on a 5\% sample with the three sampling techniques for either data set. For S2, the average activation value was not impacted as much by the tuning factor. The difference was small enough not to significantly impact the value of this metric. For S3, we saw results similar to S1. The highest values were at
$j=0$, because at this point there was the least amount of diversity in the data points; each cluster
only contributed exemplar data points. As the number of outliers increased, the distance between the distribution
became lower, the lowest point around $j~\sim 0.5$, as the tuning factor increased further and the sample contains an increasing number of outliers this value became
lower at a slower rate.

Weighted sample values are indicated by dashed lines on the \figref{fig:tf_res}. These samples
did not achieve the best value consistently and were significantly less deterministic in their performance.

\begin{table}[t]
\setlength\tabcolsep{3pt}
\begin{threeparttable}
  \begin{tabularx}{\columnwidth}{@{} L RR @{}}
  \toprule
  \textbf{Sampling Technique} & \textbf{MNIST}& \textbf{GZoo2} \\

    \midrule

  All & 9893(107) & 2097(21)  \\
  Uniform & 499(1) & 105(1)  \\
  Latent space sample & 487(9) & 108(0)\\
  Stratified by CM & 494(11)   & 105(1)  \\
  GMM (full) sample & 464(48)   & 104(4)  \\
  GMM (sph) sample & 500(11)   & 108(0)  \\
  Max margin sample & 427(85)   & 89(18)  \\
  Visually Aware Sample  & 492(8) & 492(4)\\
  EB Tree sample & 198(72) & 44(8)\\

  \bottomrule
  \end{tabularx}
    \caption{Number of data points in samples for each sampling strategy with  correctly classified (incorrectly classified)
  data points in the two data sets.}
    \tablabel{tab:sampleNumbers}
\end{threeparttable}

\end{table}

\subsection{Sample Creation Overhead}
\begin{figure}[t]
    \centering
\includegraphics[width=0.95\columnwidth]{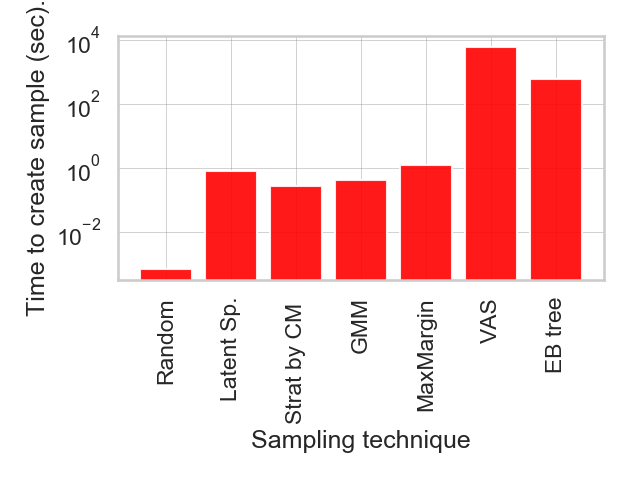}
    \vspace{-1em}
\caption{Time to create a 5\% sample for Galaxy Zoo2 dataset for all sampling techniques.} \figlabel{fig:samGen}

\end{figure}
Here, we examine the time it took to create these samples for baselines as well as for our sampling techniques.
\figref{fig:samGen} depicts the time required to generate a 5\% sample
for all sampling techniques on the Galaxy Zoo2 test data set; note the log scale on the y-axis.
Results for the MNIST data set were similar and are not shown.
The Galaxy Zoo2 test set had 2118 points, each point is a vector of size $[1, 64]$.
 Generating the uniform sample was the fastest as expected. Generating,
\textit{stratified by CM} sample, and \textit{GMM} samples, required similar time, taking less than a second. Generating
\textit{MaxMargin classification}-based sample required 1.5 seconds.  Both VAS and EB-tree samples took three orders of magnitude more time.
VAS is created with the interchange algorithm~\cite{Park16}, each point in the
data set has to be added and one point evicted, by comparing proximity of the added point with each element of the existing sample. This is $O(K_{2}N)$ where K is
the sample size and N is number of points in the data set. For large data sets as in cases of ML, the time to create this sample was unacceptably long. Boundary stitching algorithm~\cite{Wu18} is $O(NK)$.
This was faster than the VAS but still took longer than our sampling technique.

\section{Related Work}
\seclabel{relwork}
Our work is related to three different categories of research; approximate query processing, model diagnosis systems,  and model lifecycle management and
tuning systems. We review work from each of these categories below.

\subsubsection{Approximate query processing (APQ) and top-K queries} (\cite{Agarwal13,Acharya99, Babcock03, Hellerstein97}) APQ is a
well-studied area in databases and is an effective technique to deal with large-scale data.  Algorithms for exact top-k queries
are defined by the seminal work on the threshold algorithm (TA)~\cite{fagin03}, which require access to the indexed attribute(s) for a data set.
 Efficient processing of the top-k queries over samples is a challenging task~\cite{ilyas08}. Related work in this category
includes top-k processing techniques that operate on deterministic data but report
approximate answers in favor of performance. The approximate answers are usually associated with probabilistic guarantees;
indicating how far they are from the exact answer. Algorithms presented in ~\cite{theobald04} are an approximate adaptation
of TA where the approximate answers to the top-k query is associated
with probabilistic guarantees. However, like TA this algorithm requires access to sorted attributes for the underlying data.
Another approach to approximate top-k answers is considered in similarity search for multi-media databases ~\cite{amato03}. This
method uses a proximity measure to determine if a data region should be inspected. This utilizes the underlying
data distribution rather than individual column value and in that sense is closer to
our approach (i.e., instead of examining the underlying data, we utilize the latent space to create a sample).

\subsubsection{Model diagnosis systems} (\cite{Vartak18,Miao17, kahng18, Amershi15,  kahng16})
Model tracker\cite{Amershi15} is one of the earliest systems for model diagnosis. It diagnoses models
by tracking its performance using statistical measures, such as accuracy, AUC, etc. and does
not support model diagnosis for DL models. MLCube~\cite{kahng16}, one of the earlier visualization
tool for model diagnosis visualizes data from pre-computed data cubes based on features from data and model results. The data-cubes utilized
by this tool are based on less than $100$ features and like Model tracker, it pre-dates the large scale of data that must
be supported for DL model diagnosis. MISTIQUE~\cite{Vartak18} supports DL model diagnosis via examination
of model activations, their primary approach is to reduce the storage footprint required by activations.  MISTIQUE shares our goals
of reducing query runtime for model diagnosis, but it uses a different approach,
quantization and de-duplication to reduce the storage. Modelhub~\cite{Miao17} supports model diagnosis by storing learned models and training logs with an approach that reduces
storage footprint. Modelhub focuses on different artifacts, learned models and training logs, which
they store and retrieve efficiently by introducing a model versioning system and a domain-specific language for searching
through model space, solving a very different problem. DeepBase~\cite{Sellam19} supports model interpretabiltiy and diagnosis by providing a declarative abstraction to express
and execute the generation and comparision of these artifacts. DeepBase relies on the ability to encapsulate model interpretbility
questions as hypothesis functions (e.g., parts of speech tags and image captions).
DeepBase, ModelHub and MISTIQUE could benefit by leveraging our sampling techniques for their systems.
Finally, a variety of visualization tools~\cite{kahng18, Liu17, Strobelt18, tensorflow, Yosinski15} utilize activations and
gradients to interpret and diagnose DL models. All of these tools would benefit from our sampling techniques, as sampling
would help reduce the scale of data required to support model diagnosis. Activis \cite{kahng18}, for instance
selectively pre-computes values for nodes of interest to save computation and storage. Sampling techniques such as ours will
enable ML practitioners using tools such as Activis to avoid making such compromises.

\subsubsection{Model lifecycle management and model tuning} (\cite{Vartak17, mlflow, sparks15, Binnig18})  ModelDB~\cite{Vartak17} is a system
for managing of ML models and pipelines. It provides versioning and metadata-based search
and validation on models, simplifying the model building pipeline. MLflow~\cite{mlflow} tracks experiments, packages
the code to create reusable deployments and operationalizes the chosen models, addressing a very different aspect of model
lifecycle management compared to ModelDB. However, neither of these systems help manage, store, or query any DL model
diagnosis artifacts. While MLflow supports storing and tracking arbitrary artifacts in a framework and implementation agnostic manner, it does not
utilize information such as representation learned by the models to help with the selection of appropriate model. In addition
custom code has to be provided for generating and querying these artifacts in MLflow.
 These tools do not support model diagnosis or interpretability as a primary
goal, if they were to adopt model diagnosis as a goal our sampling technique could help with managing the size of data required.

\section{Conclusion and Future work}
\seclabel{conc}

Deep learning models have become an indispensable tool for a wide range of tasks, such as
image classification, object recognition, speech analysis, machine translation, and more. The task of
diagnosis for these purportedly black-box models requires additional artifacts, such as activations. These additional artifacts must be generated, stored, and queried for each DL
model being debugged. The addition of these artifacts, which can be up to three orders of magnitude larger
than the input data size for each model being diagnosed, turns the process
of building, diagnosing, and selecting DL models in to a large-scale data management challenge.
In this work, we quantify DL diagnosis workload and present a novel
sample creation technique that reduce the time and complexity required to accomplish these tasks.

The sampling technique we present in this paper focus on sampling
input data points, e.g. rows from the relation of data points and activations. The ML literature
supports the notion of reducing the number of neurons for which activations need to be calculated~\cite{Liu17,raghu17}
and queried. We would like to explore this avenue in future work.
The sampling technique described in this paper works well with supervised learning models, i.e. DL
models built with labeled data. In future work, we would like to explore our sampling technique
and their efficacy for \textit{unsupervised} DL models, such as generative models,
autoregressive models, etc.~\cite{dm} A large body of scientific data is unlabeled and requires
unsupervised learning techniques, and extending our sampling technique in this direction could
be beneficial to the scientific community working on newer data sets.

\textbf{Acknowledgements}:
This project is supported by NSF grants OAC-1739419, CCF-1535565, AST-1715122 and Charles and Lisa Simonyi Fund for Arts and Sciences, Washington Research Foundation.
\end{sloppypar}
%
\bibliographystyle{abbrv}
\bibliography{references}
\end{document}